\documentclass{llncs}
\usepackage[T1]{fontenc}
\usepackage{graphicx}
\usepackage{amsmath}
\usepackage{amssymb}
\usepackage{multirow}
\usepackage{tabularx}
\begin{document}

\title{PointNDF: Cascaded Neural Displacement Fields for Point Cloud Denoising}

\author{
Michael Y. Wang\inst{1}\orcidID{0009-0000-5005-9818} \and
Pingping Cai\inst{2}\orcidID{0000-0002-1487-1443}
}

\institute{
Dutch Fork High School, Irmo, SC 29063, USA \and
University of South Carolina, Columbia, SC 29208, USA
}

\maketitle

\begin{abstract}
Point cloud denoising aims to recover clean geometry from noisy 3D observations while preserving both fine surface details and the underlying spatial distribution of the points. Many learning-based methods adopt a two-stage pipeline that first extracts point-wise features from the noisy input and subsequently predicts point-wise displacements via simple regressions. However, these approaches often lack robust feature guidance and distribution-aware geometric refinement, causing them to fail at maintaining optimal point uniformity and geometric quality. To address these limitations, we propose \textbf{PointNDF}, a denoising framework that combines patch-level geometric context with local displacement prediction. Specifically, a Point Transformer V3 (PTv3) encoder aggregates multiscale geometric information into a global patch descriptor. Rather than directly regressing an unconstrained displacement vector, a Neural Displacement Field (NDF) fuses this global patch descriptor with per-point neighborhood features. It then predicts each displacement through bounded-vector and scale factor branches, thereby achieving a superior, more uniform point distribution. Three cascaded NDF stages propagate local features and progressively refine the point positions. Experiments on PU-Net and ModelNet10 demonstrate competitive denoising accuracy and favorable inference efficiency. The source code is publicly available at https://github.com/peridous/PointNDF.
    %The improvements are most pronounced for 50K-point clouds and higher noise levels, suggesting that the context captured by PTv3 is particularly useful for dense, strongly corrupted inputs.
    
\end{abstract}

\section{Introduction}
Point clouds provide a direct and flexible representation of 3D geometry and are widely used in autonomous systems, indoor-scene reconstruction, robotics, and digital content creation~\cite{[1],[2],[3]}. However, in practice, measurements from LiDAR, structured-light scanners, and photogrammetry pipelines are affected by sensor uncertainty, imperfect calibration, and reconstruction errors. The resulting noise and irregular sampling can obscure fine structures and reduce the reliability of downstream tasks such as surface reconstruction, recognition, rendering, and simulation~\cite{[4],[5],[6]}.

The goal of point cloud denoising is therefore not simply to smooth the input, but to move noisy points toward the underlying surface without erasing sharp features or distorting the underlying surface. Traditional approaches rely on handcrafted priors such as local surface fitting and neighborhood regularization~\cite{[7],[8]}. Although effective on smooth surfaces, these priors can oversmooth detailed geometry and may not generalize well to complex shapes.

\begin{figure}[t]
    \centering
    \includegraphics[width=1\linewidth]{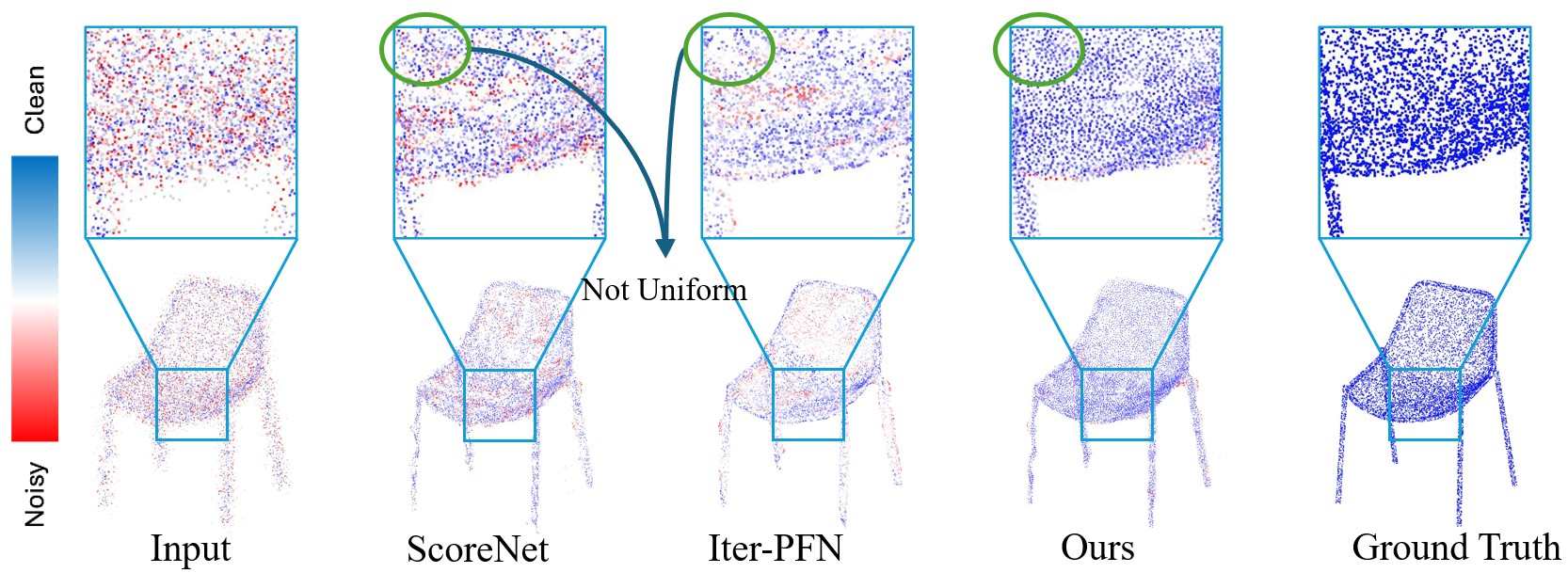}
    \caption{Visual comparison on a 10,000-point cloud corrupted by 3\% Gaussian noise. Competing methods retain visible noise or uneven local distributions, whereas PointNDF more closely recovers the reference geometry and point distribution.}
    \label{fig:teaser}
\end{figure}

Deep networks offer a data-driven alternative by learning geometric representations directly from unordered point sets~\cite{[9],[10],[11]}. Many recent methods follow a two-stage design: an encoder extracts shape features from the noisy input point cloud, and a refinement module predicts point-wise displacements toward the clean surface~\cite{[12],[13],[14],[15]}. This design exposes two closely related bottlenecks. First, features extracted from noisy coordinates must remain informative enough to represent both the local detail and the global shape. Second, the decoder must translate those features into stable point-wise displacements without introducing local point clustering or removing fine details. These challenges become more pronounced as the geometry grows denser and more complex. As illustrated in Figure~\ref{fig:teaser}, large residual noise and uneven point distributions still exist in previous methods even after denoising.

Motivated by these limitations, we investigate point cloud denoising from the perspective of improving both feature representation and geometric refinement. Our framework, PointNDF, addresses two key challenges in existing methods: effectively extracting robust geometric features from noisy observations and accurately predicting point displacements toward the underlying surface.

Specifically, we adopt Point Transformer V3 (PTv3) as the backbone encoder to capture multiscale geometric information from noisy point clouds. By aggregating these features into a global patch descriptor, PTv3 provides more robust geometric context for subsequent denoising. To improve point position refinement, we introduce a Neural Displacement Field (NDF), which combines the global patch descriptor and the per-point local shape feature to predict point-wise displacements and enables more accurate movement of noisy points toward the underlying clean surface. Furthermore, rather than applying one potentially unstable displacement, we stack three NDF modules. Each module refines the output of the preceding stage, allowing the network to remove large deviations first and make progressively smaller residual displacement vectors thereafter.

Unlike Pointfilter~\cite{[12]} and IterativePFN~\cite{[15]}, PointNDF fuses a shared PTv3 patch descriptor with per-point neighborhood features and predicts each displacement as the product of a bounded vector and a nonnegative scale factor.

We evaluate PointNDF on PU-Net and ModelNet10 at two point resolutions and three noise levels. The method achieves competitive results overall, with its clearest advantages on dense 50K-point inputs and at higher noise levels. It also offers fast inference time relative to most compared methods. These results suggest that PTv3’s patch-level context is particularly beneficial for dense point clouds to describe broader geometric relationships. In summary, the main contributions of this work are:

\begin{itemize}
    \item We adapt Point Transformer V3 to point cloud denoising and use its
    multiscale features to provide patch-level geometric context for point-wise
    refinement.

    \item We introduce a cascaded Neural Displacement Field that differs from
    direct displacement regressors by factorizing each point-wise displacement as the product of a bounded
    vector and a nonnegative scale factor.

    \item Experiments and ablations show competitive accuracy, consistent gains
    from progressive refinement, and particularly strong performance on dense
    point clouds under substantial noise.
\end{itemize}

\section{Related Work}
\textbf{Point Cloud Processing:}
Classical point cloud methods rely on local surface priors and handcrafted geometric operators~\cite{[16],[17],[18],[19],[20]}. Learning-based approaches instead process unordered points directly: PointNet~\cite{[9]} introduced permutation-invariant set encoding, PointNet++~\cite{[10]} added hierarchical local features, and DGCNN~\cite{[11]} and PointCNN~\cite{[21]} modeled local relationships through dynamic graphs and learned neighborhood transformations.

Point Transformer~\cite{[22]} and Point Transformer V2~\cite{[23]} introduced attention-based geometric modeling, while Point Transformer V3 (PTv3)~\cite{[24]} improved scalability through point serialization. Although PTv3 performs strongly on point cloud understanding, denoising additionally requires precise displacement prediction. We therefore pair its patch-level context with a local displacement decoder.

\textbf{Point Cloud Denoising:}
Early learning-based denoisers predicted point corrections or consolidation constraints, including PointCleanNet~\cite{[25]} and EC-Net~\cite{[26]}, while Total Denoising~\cite{[27]} removed the need for paired training data. Other methods reconstruct and resample a latent surface, as in DMRDenoise~\cite{[28]}, or improve sampling quality through upsampling objectives~\cite{[29]}. Probabilistic alternatives include the normalizing-flow formulation of PD-Flow~\cite{[13]}, the iterative score updates of ScoreDenoise~\cite{[14]}, and P2P-Bridge~\cite{[30]}, which uses Schrödinger bridges to formulate denoising.

Most closely related to our method are displacement-based filters. Pointfilter~\cite{[12]} uses a PCA-aligned local patch to regress direct displacements and may repeatedly apply the filter at inference. IterativePFN~\cite{[15]} instead trains graph-based IterationModules with adaptive intermediate targets. PointNDF differs by fusing a shared PTv3 patch descriptor with attention-based local features, factorizing displacements using bounded-vector and scale factor branches, and propagating local features across stages. All stages use the same symmetric Chamfer target, forming a global--local displacement field that progressively updates the entire patch.

\begin{figure}[t]
    \centering
    \makebox[\linewidth][c]{%
        \includegraphics[width=1.0\linewidth]{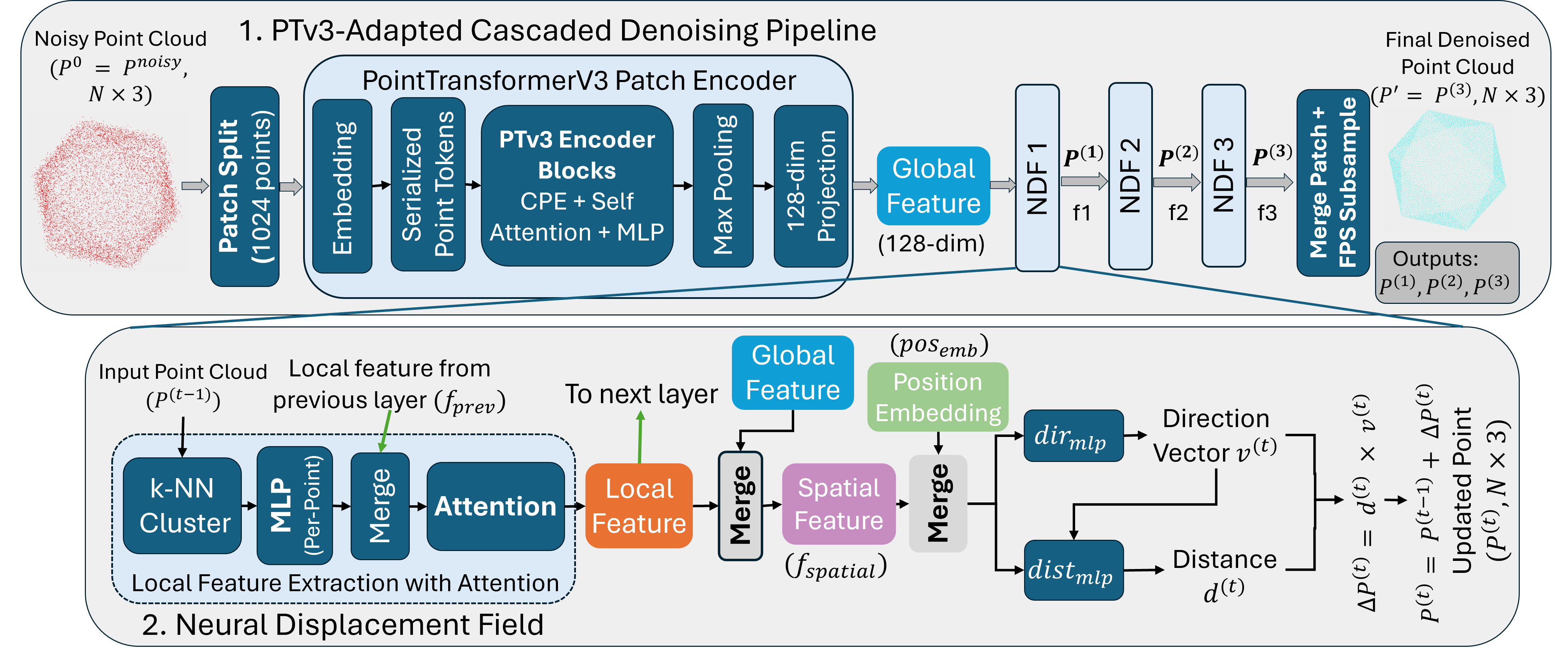}
    }
    \caption{Overview of PointNDF. PTv3 extracts a global patch descriptor that is fused with per-point local features in three cascaded Neural Displacement Field stages. Each stage predicts an additional point-wise displacement vector.}
    \label{fig:pipeline_overview}
\end{figure}

\section{Proposed Method}

\subsection{Problem Formulation}

Let $P^{\mathrm{noisy}} = \left\{p_i^{\mathrm{noisy}}\right\}_{i=1}^{N}$ denote a noisy point cloud normalized to the unit sphere. Our goal is to recover a point cloud with the same cardinality whose points lie close to the underlying clean surface. We model each observed coordinate as
\begin{equation}
    p_i^{\mathrm{noisy}} = p_i^{\mathrm{clean}} + \epsilon_i,
\end{equation}
where $p_i^{\mathrm{clean}} \in \mathbb{R}^{3}$ is the corresponding clean position from which the noisy observation is generated and $\epsilon_i \in \mathbb{R}^{3}$ represents the added noise.

Instead of generating a new point set, the network preserves the input points and predicts residual displacements through three refinement stages. Let $P^{(0)} = P^{\mathrm{noisy}}$. At stage $t$, the coordinates are updated as
\begin{equation}
 P^{(t)} = P^{(t-1)} + \Delta P^{(t)}, \qquad t=1,2,3,  
\end{equation}
where $\Delta P^{(t)}$ is the set of point-wise displacement vectors predicted at stage $t$. The final output of one network pass is $P^{\mathrm{denoised}} = P^{(3)}$.

\subsection{PTv3-Based Feature Extraction}
We use Point Transformer V3 (PTv3)~\cite{[24]} as an encoder-only feature extractor because its serialization strategy efficiently expands the spatial context available to the network. PTv3 first converts each point-cloud patch into a sparse representation using a fixed voxel grid size of $0.0066$. It then serializes the points using Z-order, transposed Z-order, Hilbert order, and transposed Hilbert order. These space-filling curves preserve local spatial proximity in the serialized sequence, enabling the attention mechanism to efficiently model geometric relationships.

The final encoder layer produces a 512-dimensional feature for each retained point. Channel-wise max pooling then summarizes the most salient responses in a patch. Specifically, let

\begin{equation}
X_b=\left\{x_{b,i}\right\}_{i=1}^{M_b},\qquad x_{b,i}\in\mathbb{R}^{512},
\end{equation}
denote the final PTv3 features for patch $b$, where $M_b$ is the number of encoded points. We compute

\begin{equation} g_b^{\mathrm{pool}} = \max_{i=1,\ldots,M_b} x_{b,i}, \qquad g_b^{\mathrm{pool}} \in \mathbb{R}^{512}. \end{equation}
Then, a linear layer projects this pooled feature to a 128-dimensional descriptor, denoted below by $g_b$. The descriptor is shared across the three refinement stages, providing a stable global reference as the point coordinates change.

%Although PTv3 produces features at multiple points, the current implementation
%uses the pooled global representation rather than directly passing pointwise
%PTv3 features to the decoder.

\subsection{Neural Displacement Field}
The global descriptor alone does not provide enough point-specific information to determine how each point should move. We therefore introduce a Neural Displacement Field (NDF) that combines global context with local geometry to predict a displacement vector for every point. The cascade contains three modules, $\mathcal{D}_1$, $\mathcal{D}_2$, and $\mathcal{D}_3$. The first operates on the noisy coordinates; the next two refine the preceding estimates. The modules share an architecture but have independent learnable parameters, allowing later stages to specialize in smaller residual displacements.

Unlike Pointfilter's local vector regressor~\cite{[12]} and IterativePFN's direct-displacement IterationModules~\cite{[15]}, PointNDF shares the PTv3 descriptor across stages, propagates local features, and uses coupled bounded-vector and scale-factor branches.

\subsubsection{Local Feature Extraction}
At each stage, the NDF first extracts a feature that describes the local surface around every current point. A point-wise multilayer perceptron (MLP) with dimensions $3 \rightarrow 16 \rightarrow 64$ embeds the current XYZ coordinates into a 64-dimensional feature. At stages two and three, this coordinate embedding is concatenated with the 64-dimensional local feature from the preceding stage. This connection propagates geometric evidence through the cascade instead of requiring each stage to infer the neighborhood structure from scratch.

Each NDF module then applies local attention over a 32-nearest-neighbor neighborhood~\cite{[24],[22]}. By jointly encoding point features and relative XYZ offsets, the attention layer captures local surface structure while remaining sensitive to the current point positions.

The resulting 64-dimensional local feature is fused with the 128-dimensional global descriptor extracted by PTv3. Specifically, the global descriptor is broadcast to all points and concatenated with each local feature:

\begin{equation}
f_{\mathrm{fuse},i}^{(t)} = \operatorname{MLP}\left([f_{\mathrm{local},i}^{(t)};\ g_b]\right),
\end{equation}
where $f_{\mathrm{local},i}^{(t)} \in \mathbb{R}^{64}$ is the local feature of point $i$ at stage $t$. A point-wise MLP with dimensions $192 \rightarrow 128 \rightarrow 128 \rightarrow 64$ maps the concatenated feature to $f_{\mathrm{fuse},i}^{(t)}$. The resulting feature preserves the local geometric detail needed for precise displacement prediction while conditioning each point’s movement on the broader shape of the patch.

\subsubsection{Bounded-Vector and Scale Factor Prediction}
Predicting a displacement as an unconstrained three-dimensional vector can entangle its orientation and magnitude. We instead predict each correction using a bounded vector and a nonnegative scalar that controls its scale. First, the positional encoding maps a 3D coordinate to a nine-dimensional feature containing its original and sinusoidal components: 
\begin{equation}
\gamma(p_i) = [p_i;\,\sin(\pi p_i);\cos(\pi p_i)].
\end{equation}
For point $i$ at stage $t$, the encoded coordinate is then concatenated with the corresponding fused feature and passed to the bounded-vector branch:
\begin{equation}
v_i^{(t)}=\tanh\!\left(\phi_{\mathrm{dir}}\left([\gamma(p_i^{(t-1)});\,f_{\mathrm{fuse},i}^{(t)}]\right)\right).
\end{equation}
The $\tanh$ activation bounds each component of $v_i^{(t)}$ in $[-1,1]$. Although $v_i^{(t)}$ is not explicitly normalized to unit length, bounding it limits the scale of the vector output.

The scale factor branch takes the encoded coordinate, the encoded bounded vector, and the fused feature as input:
\begin{figure}[ht]
    \centering
    % \captionsetup[subfigure]{font=tiny}

    % \makebox[\textwidth][c]{%
    % \begin{minipage}{1.15\textwidth}
    % \centering

    % \begin{subfigure}[t]{0.025\linewidth}
    %     \centering
    %     \includegraphics[width=\linewidth]{zoom_in_camel/clean-noise.png}
    % \end{subfigure}
    % \hfill
    % \begin{subfigure}[t]{0.11\linewidth}
    %     \centering
    %     \includegraphics[width=\linewidth]{zoom_in_camel/noisy_camel.png}
    %     \caption{Noisy}
    % \end{subfigure}
    % \hfill
    % \begin{subfigure}[t]{0.11\linewidth}
    %     \centering
    %     \includegraphics[width=\linewidth]{zoom_in_camel/score_camel.png}
    %     \caption{Score}
    % \end{subfigure}
    % \hfill
    % \begin{subfigure}[t]{0.11\linewidth}
    %     \centering
    %     \includegraphics[width=\linewidth]{zoom_in_camel/pdflow_camel.png}
    %     \caption{PD-Flow}
    % \end{subfigure}
    % \hfill
    % \begin{subfigure}[t]{0.11\linewidth}
    %     \centering
    %     \includegraphics[width=\linewidth]{zoom_in_camel/iterpfn_camel.png}
    %     \caption{Iter-PFN}
    % \end{subfigure}
    % \hfill
    % \begin{subfigure}[t]{0.11\linewidth}
    %     \centering
    %     \includegraphics[width=\linewidth]{zoom_in_camel/dmr_camel.png}
    %     \caption{DMR}
    % \end{subfigure}
    % \hfill
    % \begin{subfigure}[t]{0.11\linewidth}
    %     \centering
    %     \includegraphics[width=\linewidth]{zoom_in_camel/ptv3_camel.png}
    %     \caption{Ours}
    % \end{subfigure}
    % \hfill
    % \begin{subfigure}[t]{0.11\linewidth}
    %     \centering
    %     \includegraphics[width=\linewidth]{zoom_in_camel/clean_camel.png}
    %     \caption{Ground Truth}
    % \end{subfigure}
    % \end{minipage}%
    % }
    \includegraphics[width=1.0\linewidth]{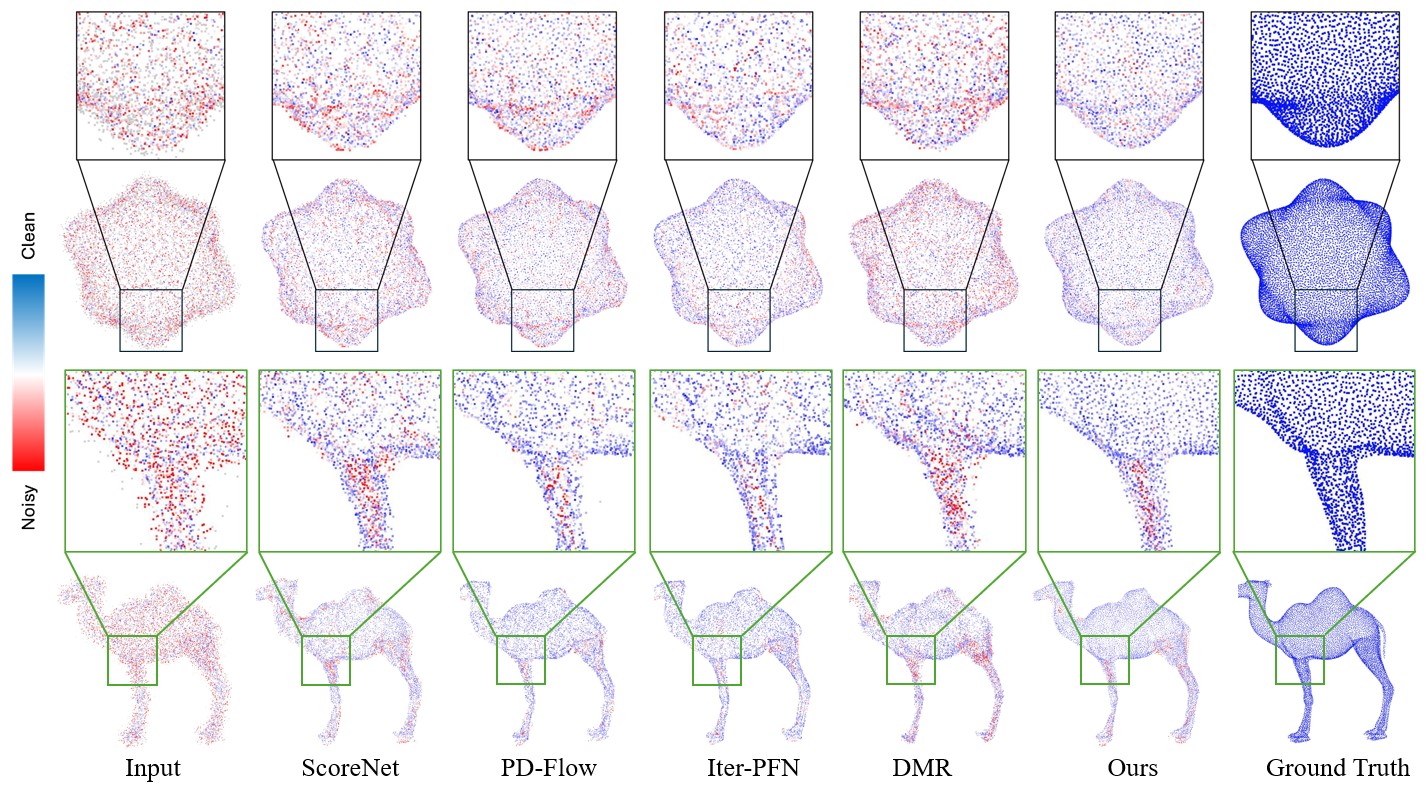}
    \caption{Qualitative comparison on 10,000-point test samples corrupted by 3\% Gaussian noise. PointNDF suppresses visible off-surface noise while preserving the overall shape and better uniformity.}
    \label{fig:qualitative_comparison}
\end{figure}
\begin{equation}
d_i^{(t)}=\left|\phi_{\mathrm{dist}}\left([\gamma(p_i^{(t-1)});\gamma(v_i^{(t)});f_{\mathrm{fuse},i}^{(t)}]\right)\right|.
\end{equation}
The absolute value makes the predicted scale factor nonnegative. The displacement and updated coordinate are then obtained:
\begin{equation}
\Delta p_i^{(t)} = d_i^{(t)}v_i^{(t)}, \quad p_i^{(t)} = p_i^{(t-1)} + \Delta p_i^{(t)}, \qquad t=1,2,3.
\end{equation}
The final denoised point cloud is $P^{\mathrm{denoised}} = P^{(3)}$. Unlike a decoder that directly emits an unconstrained three-dimensional displacement, this two-branch parameterization bounds the vector components while allowing a nonnegative scalar branch to modulate the overall correction.

\subsection{Training Objective}
During training, the three intermediate denoised point clouds $P^{(1)}$, $P^{(2)}$, and $P^{(3)}$ are supervised using the ground-truth clean point cloud $P^{\mathrm{clean}}$. The model is optimized to minimize the geometric discrepancy between the predicted and clean point positions.

We adopt Chamfer distance (CD), a nearest-neighbor point-set distance related to distance-transform methods and widely used as a reconstruction loss in 3D learning tasks~\cite{[31]}. Given a predicted point cloud $P$ and a clean point cloud $Q$, the Chamfer distance is defined as:
\begin{equation}
\mathcal{L}_{\mathrm{CD}}(P,Q) = \frac{1}{2}\left( \frac{1}{|P|}\sum_{p\in P}\min_{q\in Q}\lVert p-q\rVert_2 + \frac{1}{|Q|}\sum_{q\in Q}\min_{p\in P}\lVert q-p\rVert_2 \right).
\end{equation}
The loss computes the average nearest-neighbor distance in both directions: from the predicted point cloud to the clean point cloud and from the clean point cloud back to the prediction.
Training uses unsquared Euclidean Chamfer distance, whereas evaluation reports squared Chamfer distance.
The overall training objective is defined as:
$
\mathcal{L}=\sum_{t=1}^{3}\mathcal{L}_{\mathrm{CD}}\left(P^{(t)},P^{\mathrm{clean}}\right).
$
All three refinement stages contribute equally to the optimization objective and are matched directly to the same clean point cloud. This supervision differs from the adaptive, iteration-dependent targets used by IterativePFN~\cite{[15]}. 

\begin{table}[t]
\centering
\caption{Denoising results on the PU-Net dataset. Chamfer distance (CD, $\times 10^{-4}$) and Point-to-Mesh distance (P2M, $\times 10^{-4}$) are reported. All prediction point clouds were
evaluated using the same evaluation pipeline. The best result in each setting is shown in bold.}
\label{tab:punet_results}

\begin{tabular}{c|cccccc|cccccc}
\hline
Number of Points
& \multicolumn{6}{c|}{10K}
& \multicolumn{6}{c}{50K} \\
\hline
Noise Level
& \multicolumn{2}{c}{1\%}
& \multicolumn{2}{c}{2\%}
& \multicolumn{2}{c|}{3\%}
& \multicolumn{2}{c}{1\%}
& \multicolumn{2}{c}{2\%}
& \multicolumn{2}{c}{3\%} \\
& CD & P2M
& CD & P2M
& CD & P2M
& CD & P2M
& CD & P2M
& CD & P2M \\
\hline

DMRDenoise~\cite{[28]}
& 4.46 & 2.02
& 4.98 & 2.45
& 5.89 & 3.23
& 1.16 & 0.74
& 1.56 & 1.08
& 2.44 & 1.85 \\

PD-Flow~\cite{[13]}
& 2.13 & 0.68
& 3.26 & 1.33
& 5.20 & 2.91
& 0.65 & 0.42
& 1.42 & 1.05
& 3.90 & 3.23 \\

ScoreDenoise~\cite{[14]}
& 2.52 & 0.75
& 3.69 & 1.38
& 4.70 & 2.25
& 0.72 & 0.40
& 1.29 & 0.83
& 1.92 & 1.32 \\

IterativePFN~\cite{[15]}
& 2.06 & \textbf{0.50}
& 3.04 & \textbf{0.84}
& 4.24 & \textbf{1.69}
& 0.60 & \textbf{0.30}
& \textbf{0.80} & \textbf{0.44}
& 1.97 & 1.30 \\

P2P-Bridge~\cite{[30]}
& 2.28 & 0.69
& 3.20 & 1.11
& 3.99 & 1.76
& 0.59 & 0.33
& 0.90 & 0.58
& 1.56 & 1.12 \\

Ours
& \textbf{1.95} & 0.56
& \textbf{2.87} & 1.05
& \textbf{3.80} & 1.80
& \textbf{0.55} & 0.34
& 0.92 & 0.61
& \textbf{1.50} & \textbf{1.08} \\

\hline
\end{tabular}
\end{table}

\begin{table}[t]
\centering
\caption{Denoising results on the ModelNet10 evaluation set. Chamfer distance (CD, $\times 10^{-4}$) and Point-to-Mesh distance (P2M, $\times 10^{-4}$) are shown. The best result in each setting is shown in bold.}
\label{tab:modelnet10_results}

\begin{tabular}{c|cccccc|cccccc}
\hline
Number of Points
& \multicolumn{6}{c|}{10K}
& \multicolumn{6}{c}{50K} \\
\hline
Noise Level
& \multicolumn{2}{c}{1\%}
& \multicolumn{2}{c}{2\%}
& \multicolumn{2}{c|}{3\%}
& \multicolumn{2}{c}{1\%}
& \multicolumn{2}{c}{2\%}
& \multicolumn{2}{c}{3\%} \\
& CD & P2M & CD & P2M & CD & P2M
& CD & P2M & CD & P2M & CD & P2M \\
\hline

DMRDenoise~\cite{[28]}
& 6.52 & 4.56
& 6.74 & 4.19
& 8.04 & 5.30
& 1.77 & 1.42
& 2.18 & 1.33
& 3.15 & 2.20 \\

PD-Flow~\cite{[13]}
& 4.02 & 1.57
& 5.25 & 2.30
& 8.01 & 4.22
& 1.84 & 0.99
& 2.44 & 1.51
& 4.55 & 3.45 \\

ScoreDenoise~\cite{[14]}
& 3.90 & 1.93
& 5.40 & 2.50
& 7.89 & 3.98
& 1.46 & 0.84
& 2.18 & 1.31
& 4.26 & 2.94 \\

IterativePFN~\cite{[15]}
& \textbf{3.26} & \textbf{1.54}
& \textbf{4.52} & \textbf{1.88}
& 6.84 & \textbf{2.99}
& 1.38 & 0.69
& \textbf{1.68} & \textbf{0.83}
& 3.92 & 2.40 \\

P2P-Bridge~\cite{[30]}
& 3.84 & 1.55
& 5.44 & 2.41
& 6.88 & 3.54
& \textbf{1.31} & 0.68
& 2.08 & 1.21
& 3.55 & 2.43 \\

Ours
& 4.06 & 1.68
& 5.30 & 2.71
& \textbf{6.49} & 3.47
& 1.39 & \textbf{0.65}
& 2.10 & 1.18
& \textbf{2.97} & \textbf{1.93} \\

\hline
\end{tabular}
\end{table}
\section{Experiments}
\subsection{Comparison Methods}
We compare PointNDF with five representative learning-based denoisers: DMRDenoise~\cite{[28]}, which reconstructs a manifold representation; PD-Flow~\cite{[13]}, which uses normalizing flows; ScoreDenoise~\cite{[14]}, which iteratively follows a learned score; IterativePFN~\cite{[15]}, which performs progressive point filtering; and P2P-Bridge~\cite{[30]}, which formulates denoising using diffusion Schrödinger bridges between noisy and clean point clouds. Together, these baselines cover surface reconstruction, probabilistic modeling, iterative displacement, and diffusion-based strategies.

\subsection{Datasets}

PU-Net~\cite{[32]} contains 60 meshes, of which 40 are used for training and 20 for testing. We sample the training meshes at 10K, 30K, and 50K points and evaluate on 10K- and 50K-point clouds corrupted by Gaussian noise with standard deviations equal to 1\%, 2\%, and 3\% of the object scale.

To evaluate generalization beyond the training dataset, we also construct a fixed 20-mesh subset of ModelNet10, the ten-class subset of ModelNet introduced with 3D ShapeNets~\cite{[33]}. The evaluation subset contains two meshes from each category. These meshes are never used for training; the same shapes are sampled at 10K and 50K points and corrupted at the three PU-Net noise levels. 
\begin{table}[t]
\centering
\caption{Uniformity comparison on the PU-Net 10K test set.
Chamfer distance (CD, $\times 10^{-4}$) and uniformity
($\times 10^{-2}$) are reported. Uniformity is evaluated over
geodesic surface disks covering 0.4\%, 0.6\%, 0.8\%, and 1.0\% of the normalized mesh surface.}
\label{tab:punet_uniformity}

\begin{tabular}{c|c|c|cccc}
\hline
Method
& Noise Level
& CD
& \multicolumn{4}{c}{Uniformity ($\times 10^{-2}$)} \\
& & ($\times 10^{-4}$)
& 0.4\% & 0.6\% & 0.8\% & 1.0\% \\
\hline

DMRDenoise~\cite{[28]}
& 1\% & 4.46
& 3.31 & 4.15 & 4.93 & 5.69 \\

PD-Flow~\cite{[13]}
& 1\% & 2.13
& 0.16 & 0.21 & 0.28 & 0.35 \\

ScoreDenoise~\cite{[14]}
& 1\% & 2.52
& 0.57 & 0.73 & 0.90 & 1.09 \\

IterativePFN~\cite{[15]}
& 1\% & 2.06
& 0.26 & 0.31 & 0.36 & 0.44 \\

P2P-Bridge~\cite{[30]}
& 1\% & 2.28
& 0.40 & 0.55 & 0.71 & 0.87 \\

Ours
& 1\% & \textbf{1.95}
& \textbf{0.07} & \textbf{0.09}
& \textbf{0.13} & \textbf{0.17} \\
\hline

DMRDenoise~\cite{[28]}
& 2\% & 4.98
& 2.99 & 3.60 & 4.11 & 4.59 \\

PD-Flow~\cite{[13]}
& 2\% & 3.26
& 0.45 & 0.58 & 0.72 & 0.87 \\

ScoreDenoise~\cite{[14]}
& 2\% & 3.69
& 0.99 & 1.21 & 1.46 & 1.72 \\

IterativePFN~\cite{[15]}
& 2\% & 3.04
& 0.47 & 0.53 & 0.60 & 0.69 \\

P2P-Bridge~\cite{[30]}
& 2\% & 3.20
& 0.65 & 0.84 & 1.02 & 1.24 \\

Ours
& 2\% & \textbf{2.87}
& \textbf{0.28} & \textbf{0.39}
& \textbf{0.47} & \textbf{0.56} \\
\hline

DMRDenoise~\cite{[28]}
& 3\% & 5.89
& 3.26 & 3.79 & 4.26 & 4.73 \\

PD-Flow~\cite{[13]}
& 3\% & 5.20
& 0.74 & 0.93 & 1.13 & 1.34 \\

ScoreDenoise~\cite{[14]}
& 3\% & 4.70
& 2.16 & 2.67 & 3.19 & 3.68 \\

IterativePFN~\cite{[15]}
& 3\% & 4.24
& \textbf{0.58} & \textbf{0.58}
& \textbf{0.63} & \textbf{0.71} \\

P2P-Bridge~\cite{[30]}
& 3\% & 3.99
& 1.17 & 1.42 & 1.66 & 1.88 \\

Ours
& 3\% & \textbf{3.80}
& 0.66 & 0.87 & 1.10 & 1.29 \\
\hline
\end{tabular}
\end{table}
\subsection{Metrics}

We evaluate denoising quality and efficiency using CD, P2M, and uniformity, as well as reporting inference time separately. \textbf{Chamfer Distance (CD)} measures the similarity between the denoised and reference point clouds. The clean reference point cloud is normalized to the unit sphere, and the same center and scale transformation is applied to the corresponding prediction. For evaluation, CD is the sum of the mean squared nearest-neighbor distances in both directions. \textbf{Point-to-Mesh Distance (P2M)} measures the distance between the denoised point cloud and the ground-truth mesh. The mesh is normalized to the unit sphere, and the same transformation is applied to the prediction before computing the point-to-surface distance.
%\textbf{Average inference time} is the mean wall-clock time required to process one point cloud. Lower values are better for all three criteria.
\textbf{Uniformity Score}: We use the PU-GAN uniformity metric \cite{[29]} to evaluate the consistency of the local distribution of points within geodesic surface disks that cover 0.4\%, 0.6\%, 0.8\% and 1.0\% of the normalized mesh surface.

\begin{table}[t]
\centering
\caption{Average inference time comparison across denoising methods. Runtime is averaged over the ModelNet10 and PU-Net evaluation sets and reported in seconds per point cloud for each resolution and noise level.}
\label{tab:avg_inference_time}
\small
\begin{tabular*}{\linewidth}{@{\extracolsep{\fill}}c|ccc|ccc@{}}
\hline
Number of Points & \multicolumn{3}{c|}{10K} & \multicolumn{3}{c}{50K}\\
\hline
Noise Level & 1\% & 2\% & 3\% & 1\% & 2\% & 3\%\\
\hline
DMRDenoise~\cite{[28]} & 13.97 & 14.07 & 13.73 & 17.78 & 18.41 & 17.98 \\
PD-Flow~\cite{[13]} & 0.95 & 0.91 & 0.93 & 16.73 & 16.77 & 16.67 \\
ScoreDenoise~\cite{[14]} & 3.48 & 3.48 & 3.47 & 22.56 & 22.15 & 22.50 \\
IterativePFN~\cite{[15]} & 16.77 & 17.62 & 16.52 & 49.57 & 49.12 & 52.18 \\
P2P-Bridge~\cite{[30]} & 1.32 & 1.32 & 1.30 & 21.09 & 20.57 & 20.56 \\
Ours & 1.05 & 0.99 & 1.08 & 7.79 & 8.81 & 9.20 \\
\hline
\end{tabular*}
\end{table}
\subsection{Implementation Details}

The experiments were implemented in PyTorch and run on an NVIDIA RTX 3060 GPU. We generate qualitative visualizations with CloudCompare. The model was trained for 100 epochs with Adam, $\beta_1=0.8$, $\beta_2=0.9$, no weight decay, and a batch size of 16. The NDF decoder used a learning rate of $5 \times 10^{-4}$, whereas the PTv3 encoder used $5 \times 10^{-5}$. 
    
Training used the 40 PU-Net shapes sampled at 10K, 30K, and 50K points. Each point cloud was normalized to the unit sphere and augmented with Gaussian noise whose standard deviation was sampled from $[0.005,0.03]$, random scaling in $[0.8,1.2]$, and rotations about all three axes. Each epoch contained 12,000 dynamically generated 1,024-point patches (750 batches). To form a patch, we randomly selected a center and used its 1,024 nearest clean points as the target; noise and augmentation produced the corresponding input.

The PTv3 encoder used a grid size of $0.0066$ and produced a 128-dimensional global patch descriptor. Each NDF stage used a 32-nearest-neighbor local transformer. The three NDF refinement stages were trained jointly using equally weighted symmetric, unsquared CD losses. Reported CD values use the squared, unit-sphere-normalized definition described above. The complete model contained 39.1 million trainable parameters.

During inference, farthest-point sampling selected 30 input-patch centers for 10K point clouds and 120 for 50K point clouds. Each 1,024-point patch was processed independently, after which the patch outputs were concatenated and downsampled by farthest-point sampling to restore the original point count. 

All methods were evaluated using the same normalized inputs and output evaluation pipeline. Runtime was measured on the same NVIDIA RTX 3060 GPU using each method's recommended inference settings.

\subsection{Results and Discussion}

Tables~\ref{tab:punet_results} and~\ref{tab:modelnet10_results} summarize the
quantitative results. On PU-Net, PointNDF obtains the lowest CD in five of the six settings and the best P2M at 50K with 3\% noise. IterativePFN produces lower P2M in the remaining PU-Net settings, showing that the two metrics capture different aspects of reconstruction: CD rewards agreement between the two sampled point sets, whereas P2M directly measures proximity to the continuous surface.

On ModelNet10, PointNDF is less competitive at 10K, where IterativePFN leads at the lower noise levels. Its relative performance improves at 50K: it achieves the lowest P2M at 1\% noise and the lowest CD and P2M at 3\% noise. Because ModelNet10 is not used for training, the strong 50K results provide evidence of cross-dataset generalization when dense sampling supplies sufficient context and the corruption is large enough to benefit from global reasoning. Figure~\ref{fig:qualitative_comparison} supports this interpretation: the reconstructed shape remains coherent while visible off-surface noise is reduced.

Table~\ref{tab:punet_uniformity} compares the local uniformity distributions of the PU-Net 10K outputs. PointNDF achieves the lowest uniformity error across all four disk sizes at 1\% and 2\% noise. At 3\% noise, IterativePFN produces the most uniform distribution, while PointNDF retains the lowest CD.

Table~\ref{tab:avg_inference_time} shows that PointNDF's competitive accuracy also offers favorable efficiency. At 10K points, PointNDF runs in about one second per cloud, while at 50K, it is the fastest evaluated method in all three noise settings. Overall, PointNDF is not uniformly best across every resolution, noise level, and metric. Its relative strength at higher resolution and under stronger noise nevertheless supports the intended role of PTv3: using broader geometric context to guide difficult point-wise displacement predictions.
\begin{table}[t]
\centering
\caption{Backbone ablation comparing PTv3 and PointNet++ within PointNDF.
Values are averaged across 1\%, 2\%, and 3\% Gaussian noise levels.
CD and P2M are reported in $\times 10^{-4}$.}
\label{tab:pointnet_compare}

\begin{tabular}{c|c|c|c|c}
\hline
Dataset & Metric & PN++ & PTv3 & Error Reduction \\
\hline
\multirow{2}{*}{PU-Net 10K}
& CD  & 3.00 & 2.87 & +4.3\% \\
& P2M & 1.22 & 1.14 & +6.7\% \\
\hline

\multirow{2}{*}{PU-Net 50K}
& CD  & 1.15 & 0.99 & +13.9\% \\
& P2M & 0.82 & 0.68 & +16.9\% \\
\hline

\multirow{2}{*}{ModelNet10 10K}
& CD  & 5.66 & 5.28 & +6.6\% \\
& P2M & 2.68 & 2.62 & +2.4\% \\
\hline

\multirow{2}{*}{ModelNet10 50K}
& CD  & 2.98 & 2.15 & +27.7\% \\
& P2M & 1.51 & 1.25 & +16.8\% \\
\hline
\end{tabular}
\end{table}

\subsection{Ablation Studies}

\begin{table}[t]
\centering
\caption{Stage-depth ablation on the PU-Net dataset. Chamfer distance
(CD, $\times 10^{-4}$) is reported for independently trained PointNDF
models containing one, two, or three NDF refinement stages.}
\label{tab:cascade_ablation}

\renewcommand{\arraystretch}{1.15}
\footnotesize

\begin{tabular}{c|ccc|ccc}
\hline
\multirow{2}{*}{NDF Stages}
& \multicolumn{3}{c|}{10K}
& \multicolumn{3}{c}{50K} \\
& 1\% & 2\% & 3\%
& 1\% & 2\% & 3\% \\
\hline
1
& 2.204 & 3.488 & 4.870
& 0.674 & 1.311 & 2.297 \\

2
& 2.009 & 2.989 & 3.980
& 0.575 & 0.974 & 1.607 \\

3
& \textbf{1.954} & \textbf{2.865} & \textbf{3.802}
& \textbf{0.550} & \textbf{0.916} & \textbf{1.501} \\
\hline
\end{tabular}

\end{table}

\begin{table}[t]
\centering
\caption{Displacement-parameterization ablation on PU-Net.
Chamfer distance (CD, $\times 10^{-4}$) is reported at each noise level.}
\label{tab:decoder_ablation}
\small
\setlength{\tabcolsep}{4pt}

\begin{tabular}{l|ccc|ccc}
\hline
& \multicolumn{3}{c|}{10K}
& \multicolumn{3}{c}{50K} \\
Parameterization
& 1\% & 2\% & 3\%
& 1\% & 2\% & 3\% \\
\hline
Direct XYZ displacement
& 2.005 & 2.952 & 3.860
& 0.565 & 0.953 & 1.612 \\

Unit direction $\times$ step size
& 2.017 & 2.981 & 3.995
& 0.568 & 0.974 & 1.663 \\

Bounded vector $\times$ scale factor (Ours)
& \textbf{1.954} & \textbf{2.865} & \textbf{3.802}
& \textbf{0.550} & \textbf{0.916} & \textbf{1.501} \\
\hline
\end{tabular}
\end{table}

We first compare PTv3 with PointNet++ while keeping the decoder and training protocol unchanged. As shown in Table~\ref{tab:pointnet_compare}, PTv3 improves all averaged CD and P2M results, with the largest gains at 50K, supporting the benefit of richer patch-level context for dense point clouds. The improvement is especially pronounced on ModelNet10 at 50K, where PTv3 reduces CD by 27.7\% and P2M by 16.8\%. 

We next evaluate cascade depth using independently trained one-, two-, and three-stage models. Table~\ref{tab:cascade_ablation} shows consistent CD reductions as depth increases, with the largest gains at higher noise levels. At 3\% noise, the three-stage model reduces CD by 21.9\% at 10K and 34.7\% at 50K compared with the one-stage model. 

Finally, Table~\ref{tab:decoder_ablation} compares direct XYZ regression, unit-direction prediction, and the proposed bounded-vector and scale factor design. Our parameterization achieves the lowest CD across all six settings, while the unit-direction variant performs worst throughout. Compared with direct XYZ regression, the proposed design reduces average CD by 2.2\% at 10K and 5.3\% at 50K. 

\section{Conclusion}

We presented PointNDF, a point cloud denoiser that combines a PTv3 patch encoder with three cascaded Neural Displacement Field modules. The design separates patch-level geometric context from local displacement prediction: PTv3 provides a stable patch-level descriptor, while each NDF stage predicts a point-wise displacement vector for every point. Experiments on PU-Net and the unseen ModelNet10 subset show competitive accuracy and favorable inference efficiency, with the strongest results on dense 50K-point clouds and under higher noise. Backbone, cascade-depth, and displacement-parameterization ablations support the three main design choices, showing larger PTv3 gains at high resolution, monotonic improvements with additional refinement stages, and consistent benefits from the bounded-vector and scale-factor parameterization. The method remains less consistent on sparse, low-noise inputs, where local denoisers can perform better. Future work should therefore investigate adaptive numbers of refinement stages, stronger surface-aware objectives, and broader training data to improve performance across sampling densities without sacrificing scalability.

 % PointNDF bibliography in Springer LNCS-style formatting.
% Prepared from the 33 references in the supplied paper source.
% Citation keys are retained exactly (e.g., \cite{[1]}).
% This bibliography is hand-formatted, not generated by BibTeX.

\end{document}